\def\year{2022}\relax
\documentclass[letterpaper]{article} 
\usepackage{aaai22}  
\usepackage{times}  
\usepackage{helvet}  
\usepackage{courier}  
\usepackage[hyphens]{url}  
\usepackage{graphicx} 
\usepackage{natbib}  
\usepackage{caption} 
\DeclareCaptionStyle{ruled}{labelfont=normalfont,labelsep=colon,strut=off} 
\usepackage{algorithm}
\usepackage{algorithmic}

\usepackage{enumitem}
\usepackage{kotex}
\usepackage{dialogue}

\usepackage{newfloat}
\usepackage{listings}
\floatstyle{ruled}
\newfloat{listing}{tb}{lst}{}
\floatname{listing}{Listing}
\title{When Crowd Meets Persona: \\Creating a Large-Scale Open-Domain Persona Dialogue Corpus}
\author{
    Won Ik Cho\equalcontrib\textsuperscript{\rm 1}\thanks{Work done before graduation.}, Yoon Kyung Lee\equalcontrib\textsuperscript{\rm 2}, Seoyeon Bae\textsuperscript{\rm 2}, Jihwan Kim\textsuperscript{\rm 1}\textsuperscript{$\dagger$},\\ Sangah Park\textsuperscript{\rm 3}\thanks{Work while at DeepNatural AI.}, Moosung Kim\textsuperscript{\rm 4}, Sowon Hahn\textsuperscript{\rm 2}, Nam Soo Kim\textsuperscript{\rm 1}
}
\affiliations{
    \textsuperscript{\rm 1} Department of Electrical and Computer Engineering, Seoul National University, Korea\\
    \textsuperscript{\rm 2} Department of Psychology, Seoul National University, Korea\\
    \textsuperscript{\rm 3} DeepNatural AI, Seoul, Korea 
    \textsuperscript{\rm 4} Smilegate AI, Pangyo, Korea

}

\usepackage{bibentry}

\begin{document}

\maketitle

\begin{abstract}
    Building a natural language dataset requires caution since word semantics is vulnerable to subtle text change or the definition of the annotated concept. Such a tendency can be seen in generative tasks like question-answering and dialogue generation and also in tasks that create a categorization-based corpus, like topic classification or sentiment analysis. Open-domain conversations involve two or more crowdworkers freely conversing about any topic, and collecting such data is particularly difficult for two reasons: 1) the dataset should be ``crafted" rather than ``obtained" due to privacy concerns, and 2) paid creation of such dialogues may differ from how crowdworkers behave in real-world settings.
    In this study, we tackle these issues when creating a large-scale open-domain persona dialogue corpus, where persona implies that the conversation is performed by several actors with a fixed persona and user-side workers from an unspecified crowd.
\end{abstract}

\section{Introduction}

Modern conversational agents frequently have a specific persona designed so that users perceive them as having human-like characteristics \cite{seeger2021texting}. Cognitive and emotional characteristics such as personality, long-term memory, and an empathic character are also taken into account during the design process to make the user feel more engaged with these agents, thereby resulting in longer conversations \cite{schuetzler2018influence}.

Agents with specific persona often face challenging circumstances where they need to remember the dialogue history and maintain a consistent  persona to answer queries \cite{xu2021beyond}. For instance, if the agent is described as a high school boy who likes playing piano, conflict may arise if the user asks him about his experience as a career woman. Also, the agent's attitude towards the conversation partner is usually expected to remain consistent within a dialogue. If the conversation has been continued in a favorable manner, a sudden turn to a cold atmosphere may cause awkwardness. All of these factors contribute to a fluent conversation flow.

Generally, various heuristics are required to let the agent continue the conversation fluently, for instance to question understanding, to generate an answer, to engage in chit-chat, or for bypassing, or fallback instruction \cite{mctear2018conversational}. These heuristics relate to the functional aspect of the conversational agent, instead of the agent's cognitive and psychological aspects, such as the maintenance of `self' or the prevention of contradiction with previous discourses \cite{shuster2021me}. In this regard, we deemed it beneficial for both implementation and analysis of conversational agents to make up a dialogue dataset where the agent remembers itself and is consistent throughout conversation. However, constructing the dialogue dataset from scratch is not as straightforward as collecting our usual conversation.

\subsection{Why Creating a Dialogue Dataset Is Difficult}

Creating a dialogue dataset with two or more participants is a challenging process because a successful conversation between two participants requires several conditions, such as checking for common ground \cite{stalnaker2002common}, forming rapport \cite{cassell2007coordination}, and aligning communication style \cite{tannen2005conversational}. The construction becomes much more complicated when it comes to an open domain dialogue without a specific topic since the participants' common ground and interests usually differ, leading to rapid termination of the conversation.

It is difficult to use strategies adopted in task-oriented dialogues such as manual-based conversation (Wizard-of-Oz) \cite{wen2017network} or self-play \cite{shah2018building}. Manuals may not cover the variety of topics that can appear in open-domain dialogues \cite{godfrey1992switchboard,li2017dailydialog,zhang2018personalizing}, and self-play may face the limitations of content and naturalness. This inevitably calls for the participation of multiple speakers in the construction of open-domain dialogues \cite{dinan2018wizard,rashkin2019towards,roller2021recipes,xu2021beyond}.

\subsection{Open-domain Persona Dialogue}

If multiple speakers take part in a conversation, the usual dialogue continues in the way that the participants talk about daily life topics to exchange information or to enhance social bonding \cite{li2017dailydialog}. Such a setting assigns equal obligation to participants, with both struggling to lead the conversation so that the dialogue does not fall into an awkward atmosphere that lacks meaningful interaction. However, if one participant is assigned an obligation to have a conversation with those who first talk to them, just like conversational agents in service or non-playing characters of online games, the relationship and role of speakers may vary from the default setting. We regard this as a special form of persona dialogue.

In the literature, persona dialogues have been examined and crafted as representative open-domain interactions \cite{roller2020open}. In this regard, multiple persona dialogue datasets with a dual-play approach amongst participants have been created. Various persona dialogue datasets have been constructed with dual-play strategy between participants \cite{zhang2018personalizing}. While `persona' is generally accepted as a long-term social characteristic of a speaker, most studies concentrated on setting up a specific situation, with a few attributes, at the beginning of each dialogue and then asked participants to adhere to the given condition in the dialogue phase \cite{zhang2018personalizing,xu2021beyond}. This brings a large number of persona conditions and conversations that can be followed, but does not necessarily cover the long-term social characteristics that the other participant may feel rapport or friendliness. We deemed that another form of role-playing is necessary for the creation of a persona dialogue dataset with long-term social characteristics. At the same time, participants' rudeness or offense toward each other should be prevented even if they feel friendly toward each other; given that such rapport is created from an artificial situation, participants should be safe from potential mental harm that originates from human interaction.

\subsection{Our Study}

Our study focuses on creating a large-scale, open-domain persona dialogue dataset for Korean, guaranteeing a safe and non-superficial conversation between the participants. 
Our study is three-fold: 

\paragraph{Setting} We first assume the setting in which persona participants talk with user participants \cite{zhang2018personalizing,roller2020open} and where users first initiate the conversation, given the detail of persona profiles. We prepare conversation guidelines for both sides, with more obligation and the leading role assigned to the persona side. Persona participants should go through an interview to be hired as an actor and participate in the conversation.

\paragraph{Collection} After all the settings, we advertise among the worker community of a crowdsourcing platform to recruit user participants who are interested in talking with actors. Conversations are initiated with the users' message, and a web application is constructed to let participants have a conversation--and at the same time to allow the manager of the crowdsourcing platform to check on conversations for persona-user moderation.

\paragraph{Analysis} After collecting each conversation, we let the participants fill out a survey form that asks about their satisfaction with the conversation, along with details including fun, friendliness, and connectedness. We also interview with the actors and the moderator to identify their thoughts about the conversation procedure, where they felt fun or experienced difficulties. This process provides supporting details for our strategy and makes our scheme sustainable.

\medskip
With these settings, collection schemes and analyses, we aim to show that our strategy helps make up a large-scale, open-domain persona dialogue corpus with a small group of actors and crowd-user participants, providing both groups with a satisfactory experience in a talk-to-earn paradigm. We raise three research questions here:

\begin{itemize}
    \item \textbf{RQ 1:} What should be considered in accommodating the construction of a successful dialogue dataset? Which factors are important for persona dialogues and which make the process challenging?
    \item \textbf{RQ 2:} What is the role of the moderator in large-scale dialogue dataset construction? How is this different from other natural language process datasets?
    \item \textbf{RQ 3:} Will the above considerations help reach an intended construction process and output? Will they guarantee a satisfactory experience for participants while simultaneously preserving diverse persona characteristics in the corpus?
\end{itemize}

\section{Related Work}

\subsection{Conversation Datasets}

Dialogue has long been a topic of interest, and its construction schemes have changed a lot as well. Early approaches excerpted scripts from real conversation with minimal cleansing, while recent approaches build dialogue dataset from scratch and add some relevant tags on it.

\paragraph{Switchboard}

Switchboard \cite{godfrey1992switchboard} is an early multi-speaker corpus of conversational speech and text, made up of telephone conversations on pre-specified topics. It includes 2,500 conversations by 500 speakers from around the United States, collected automatically over T1 lines at Texas Instruments. Also, words for each recording are time-aligned for word transcription. Further studies such as dialogue act annotation and pragmatic analyses have been performed based on this, but it has passed about 30 years since the publication, making the dataset a bit outdated. Also, whether or not users of the telephone service have provided consent, collecting the random conversation from lines makes the dataset vulnerable to privacy and license issues.

\paragraph{Dailydialogue}

Dailydialog \cite{li2017dailydialog} is a collection of human-written and less noisy, multi-turn dialogues. The dataset reflects daily communication way and covers various topics about daily lives, including school life, work, tourism, relationships, etc. Also, emotion and dialogue acts are labeled along with the utterances, giving rich information on the analysis of the dataset. However, the raw data was crawled from various websites, which serve for an English learner to practice English dialogue in daily life, making the dataset more formal than other social media-crawled datasets. Also, dialogues do not necessarily reflect the personal characteristics of each speaker, making most of the dialogues similar to each other. In this setting, it is difficult to comprehend if the speakers are engaged with each other or feeling fun or empathy during the conversation, since the lack of shown persona may lead to the absence of rapport that yields superficial content.

\subsection{Work on Open-domain Persona dialogue}
The free chat generated by open-domain conversation agents is increasingly becoming coherent and engaging. Existing models, however, fail to generate lengthy talks (no more than five turns), merely repeat the user's remarks, and forget what the user said previously \cite{xu2021beyond}.
It's also well-known that generative models hallucinate knowledge by generating erroneous information and are incapable of interpreting external knowledge outside their limiting model parameters \cite{roller2021recipes}. Recent initiatives include role-playing games involving human crowd workers \cite{shuster2021me}, the creation of more than one thousand sets of personas \cite{zhang2018personalizing}, the maintenance of lengthy conversations \cite{xu2021beyond}, and empathetic response \cite{rashkin2019towards,smith2020improving}.

\paragraph{PersonaChat}
The PersonaChat dataset \cite{zhang2018personalizing} is a collection of personal chit-chat dialogues. Each pair of speakers bases their conversation on a specific persona profile. The dataset is divided into three stages: personas, revised personas, and persona chat. One persona is expressed in five sentences, and each sentence is rewritten to avoid redundancy. Following that, two pairs of crowdsourcing workers generated data by talking while acting out a specific persona (over 160k utterances).


\paragraph{Empathetic Dialogues and Wizard of Wikipedia}
The Empathetic Dialogues \cite{rashkin2019towards} comprises dialogue data featuring empathetic expressions. Listeners respond with empathy when they hear a speaker use a specific emotive term in a specific situation. The Wizard of Wikipedia \cite{dinan2018wizard} comprises scraped text from Wikipedia pages. Workers in expert roles spoke on multiple topics and associated each utterance with a Wikipedia sentence. The model trained on these data, therefore, can use knowledge from the Internet into free-chat. The key to open domain chatbots is to maintain engaging and lengthy dialogues. The model that contained all three components generated a more natural response, although qualitative markers were still poor. It becomes difficult to satisfy users with conversational agents that have natural, coherent conversations. It is currently insufficient to conclude that high scores for user engagement and human-likeness are indicative of user satisfaction. Therefore, higher-level concepts such as the personality and empathy qualities of an interactive agent should be properly specified in order to maintain intriguing and engaging interactions.

\paragraph{Multi-Session Chat}
Multi-Session Chat \cite{xu2021beyond} is a collection of multi-session chats between human crowd workers, in which the conversation was interrupted and restarted for as little as 7 hours and as long as seven days, as compared to the previous brief conversations that lasted less than an hour. The shorter the discussion learned by the existing open-domain dialogue models, the longer the conversation could not be continued and the memory was lost. For this reason, models that can hold a longer context are required. In order to recall what was discussed in prior sessions, a brief recap of dialogues in each episode was also labeled. Thus, another crowdworker who follow-up would be able to readily recollect the preceding discourse. A generative model trained with the aforementioned dataset was able to generate more coherent utterances than previous models. However, training only the recommended long-term memory and summary did not simply raise the user's score for engagingness. 
What aspects of conversations appeal to or intrigue users need extensive examination. 

\section{Method}

Our data construction procedure consists of four steps:  \textit{conversation guideline, pilot study, actor}\footnote{We do not specify whether the participant is an actor or an actress, but here, the word \textit{actor} is used to unify both terms, and there is no intention of the bias on biological gender.} \textit{recruitment,} and  \textit{dialogue collection}\footnote{There is a subtle difference between dialogue and conversation, but here the two words are used interchangeably.}.   

\subsection{Conversation Guideline}

Let \textit{persona} be persona participants and \textit{user} be crowdworking user participants. Conversation guidelines are documented for both the persona side and user side. The overall organization of both sides is similar, but the persona side includes more specifications.

\subsubsection{Profile}

\begin{figure}[t]
\centering
\includegraphics[width=0.8\columnwidth]{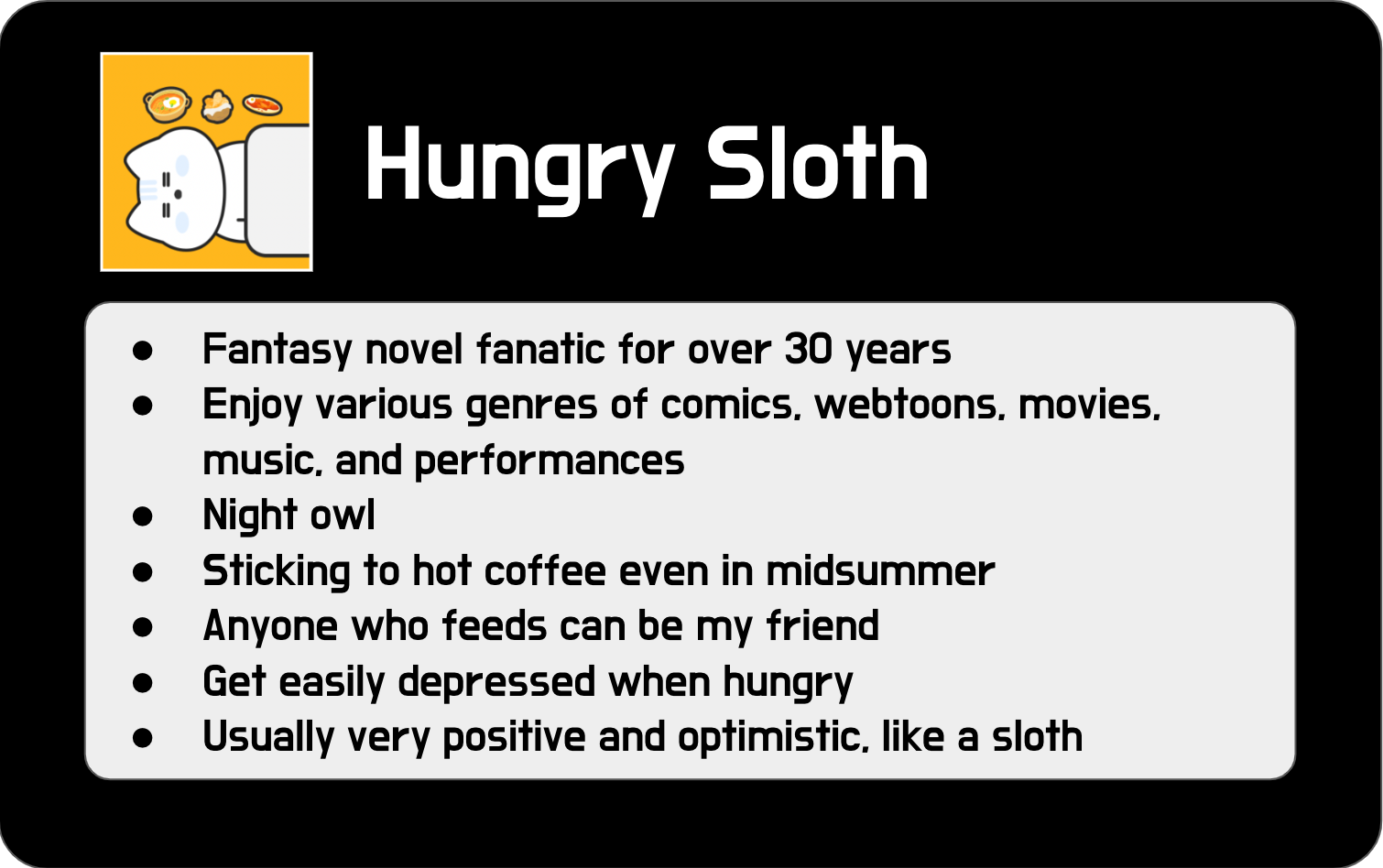} 
\caption{Example of the persona profile.}.
\label{fig:profile}
\end{figure}

The Persona side necessitates a profile that can represent the participant, which covers a slightly wider area compared to the concept of persona in the literature \cite{zhang2018personalizing,roller2020open}. A profile includes the name of the persona, the picture, and more than three information pieces on the persona. At this time, the persona can be chosen regardless of its animacy or humanness, but the participant must prepare a profile that s/he can converse while preserving the persona's consistency. In addition, the profile should NOT incorporate content that can specify the participant's personal information or induce political or social controversy. The profile picture should depict the persona appropriately, but material that violates license policies should not be used. The example of a persona profile (\textit{Hungry Sloth}) is available in Figure~\ref{fig:profile}.

Profiles prepared by personas are announced to the user side, and users decide whether to have a conversation with each of them based on the profile. Users who decide to talk with a persona should start a conversation first, and the persona has an obligation to proceed conversation with such users.

\subsubsection{Do's and Dont's}

In conversation, users and personas are given similar instructions. Typical do’s and dont’s for both sides are as follows, and some are inspired by the maxims of conversation \cite{LogicandConversation}.
All participants SHOULD:

\begin{itemize}
    \item Be respectful in the conversation. Please remember that whom you are talking with is another being who has emotion and thoughts.
    \item Be polite to each other, regardless of the usage of honorifics\footnote{Important issue in the Korean language where the social relationship affects the syntactic components of utterances \cite{Byon+2006+247+276}, but this might refer to formality in other languages without such syntactic components.}. 
    \item Be sincere in the conversation. Remember and remind the context of the dialogue as much as possible.
    \item Be cooperative in the conversation. Talk about topics that can be interesting or satisfying for both of the participants.
\end{itemize}
Also, the participants should NOT:
\begin{itemize}
    \item Ask for the other’s identity unless they reveal it. Remember that the conversation is being taken remote and in text, so that one may not want to reveal their demographic status or personal information.
    \item Include sensitive information (self’s or other’s personal information, political stance etc.) in the conversation. Please remember that such content may cause harm if the corpus is distributed publicly or used as training data.
    \item Utter hate speech that contains social bias or toxic expressions. Social bias denotes judging people depending on their identity as a minority or underrepresented group \cite{waseem2016hateful}, and toxic expressions include insults, sexual harassment, and offensive languages such as sarcasm and unethical expressions\footnote{We referred to PyCon KR Code of Conduct \url{https ://pycon.kr/2020/about/coc/} for further guideline.} \cite{davidson2017automated,moon2020beep}.
\end{itemize}

\subsubsection{Conversation Strategies}

To help lead the fluent conversation, special recommendations were provided to personas. Strategies below are a variety of skills they can utilize in awkward and irritating situations or when the conversation does not continue smoothly.

\begin{itemize}
    \item To the extent that it does not reveal the persona's personal information or identity, fully utilize the situation the persona is going through.
    \item Skim the conversation history with the user and think about the storyline to proceed the dialogue with.
    \item Consider which kind of topic the user will actively react to.
    \item If established a persona profile of a specific profession (e.g., professor, actor, director, etc.), keep the conversation using background information or commonsense knowledge.
    \item Once set up a unique persona, make sure not to run out of content or deter the tension of the conversation by appealing about it too hastily.
\end{itemize}

In addition, though personas should follow up the conversation of the user side, they were instructed to utter a lot of their own stories as well. This is to check if they maintain the persona consistently, and it is a process of explicitly or implicitly inserting contents about themselves into a dialogue.

\subsubsection{Miscellaneous}

The following instructions are given to the persona participants, who are assigned far more obligations to lead the conversation.
\begin{itemize}
    \item Take part in a conversation in view of persona, not the self in the real world.
    \item The action or conversation should at least adjust to the persona profile which the participant has decided and notified before the conversation.
    \item If the message from the user is sparse or has stopped, send a message first from time to time\footnote{This is denoted as `\textit{seon-tok}, 선톡’ in Korean, which reflects the speaker’s (or sender's) interest towards addressee (receiver).}. The recommendation is once in three times.
\end{itemize}

To facilitate conversation with personas, some instructions are given to user participants as well. 

\begin{itemize}
    \item Specific celebrities, brands, or artworks (movie, music, etc.) can be mentioned unless they do not incorporate severe intention of criticism or advertisement.
    \item Do not lead the dialogue with a topic that makes it difficult to sustain the tension.
    \item Insult or harassment is not tolerated under any circumstances, and expressions that could be misunderstood as profanity terms, even if not actually offensive, should not be used.
\end{itemize}

The above strategies and instructions apply to general conversation guidelines, independent of the user pool or crowdsourcing platform to work with. Additional instructions for persona actors and user-side workers, regarding participation time and restrictions in chatting, are introduced in the main dialogue collection phase.


\subsection{Pilot Study}

One of the goals of this study is to discern the characteristics of a conversation taken between a specific persona and the user. Therefore, in the pilot study, we verified the appropriateness of the conversation guideline with a small amount of conversation by limited participants.

\subsubsection{Participants}

For the pilot study, three persona (hereafter \textit{actors}) and five user (hereafter \textit{workers}) participants\footnote{The term `participant' is used to denote the group of people who take part in the experiment. This term is used interchangeably with `actor' or `worker' in the main construction phase, who are recruited, hired, or crowdsourced, for scalable corpus construction.} were invited, mainly of researchers and undergraduates who are interested in the project. Each actor prepared a profile and was equipped with the instructions by having a conversation with the authors. The conversation manual was provided to workers as well.

\subsubsection{Conversation}

We utilized a publicly available messenger for the pilot study. We used Kakaotalk\footnote{\url{https://play.google.com/store/apps/details?id=com.kakao.talk}} provided by the Kakao Corp., which is a mobile and desktop app familiar to most Korean citizens and provides an anonymous chat service. 
In detail, in \textit{open profile chatting}, one can set a profile for anonymous chatting, preparing the name, the picture, and a short profile. The profile created by each actor is announced to the user side, and workers start a conversation based on it.

12,000 per dialogue was paid as compensation for actors, while workers participated voluntarily based on their interests. The boundary of the dialogue was defined as the point at which the dialogue has explicitly ended and moved on to the start point of the next dialogue.

\subsubsection{Survey}

Workers conduct a score-based survey related to friendliness, likeability, connectedness, and interestingness at the end of each conversation\footnote{Note that this survey process differs from the survey that is conducted in the main dialogue collection.}. A total of 20 dialogues were conducted, and the following findings were obtained through analysis with the above survey.

\begin{itemize}
    \item Workers find a dialogue or persona likeable if there is an eye-catching concept displayed well in the conversation.
    \item Workers find a dialogue or persona likeable if a substantial amount of empathy is exhibited in the conversation, such as sharing experience, appreciation, agreement etc.
    \item A balance between self-disclosure, question, and empathy is important for fluent conversation.
    \item High-level consistency and significant characteristics of persona do not necessarily guarantee interestingness and likeableness.
\end{itemize}

\subsection{Actor Recruitment}

In actor recruitment, we gather persona participants for the main dialogue collection. In this process, 20 actors applied in, and 11 were selected based on interviews.

At this stage, for scalable and reproducible data collection, a \textit{moderator} appears and links workers and researchers (authors). The moderator equals a manager of a crowdsourcing platform with a human resource pool from which participants are recruited. S/he helps with the recruitment of participants and communication with them, at the same time managing the sourcing work and the worker education \cite{yang2022apeach}.

In detail, the moderator interviews each persona applicants and the researchers make a decision based on the interview data. The following criteria are considered in the actor selection:

\begin{itemize}
    \item Is the Persona likeable, or does the persona concept remain consistent?
    \item Shows a nice ping-pong in conversation and fluently utilizes echoing or counter-questions?
    \item Has the ability to lead a conversation that is not loose (dense) for a long period?
    \item Is self-disclosure dense and natural?
    \item Is good at being empathetic or sincere?
\end{itemize}

\begin{figure}[t]
\centering
\includegraphics[width=\columnwidth]{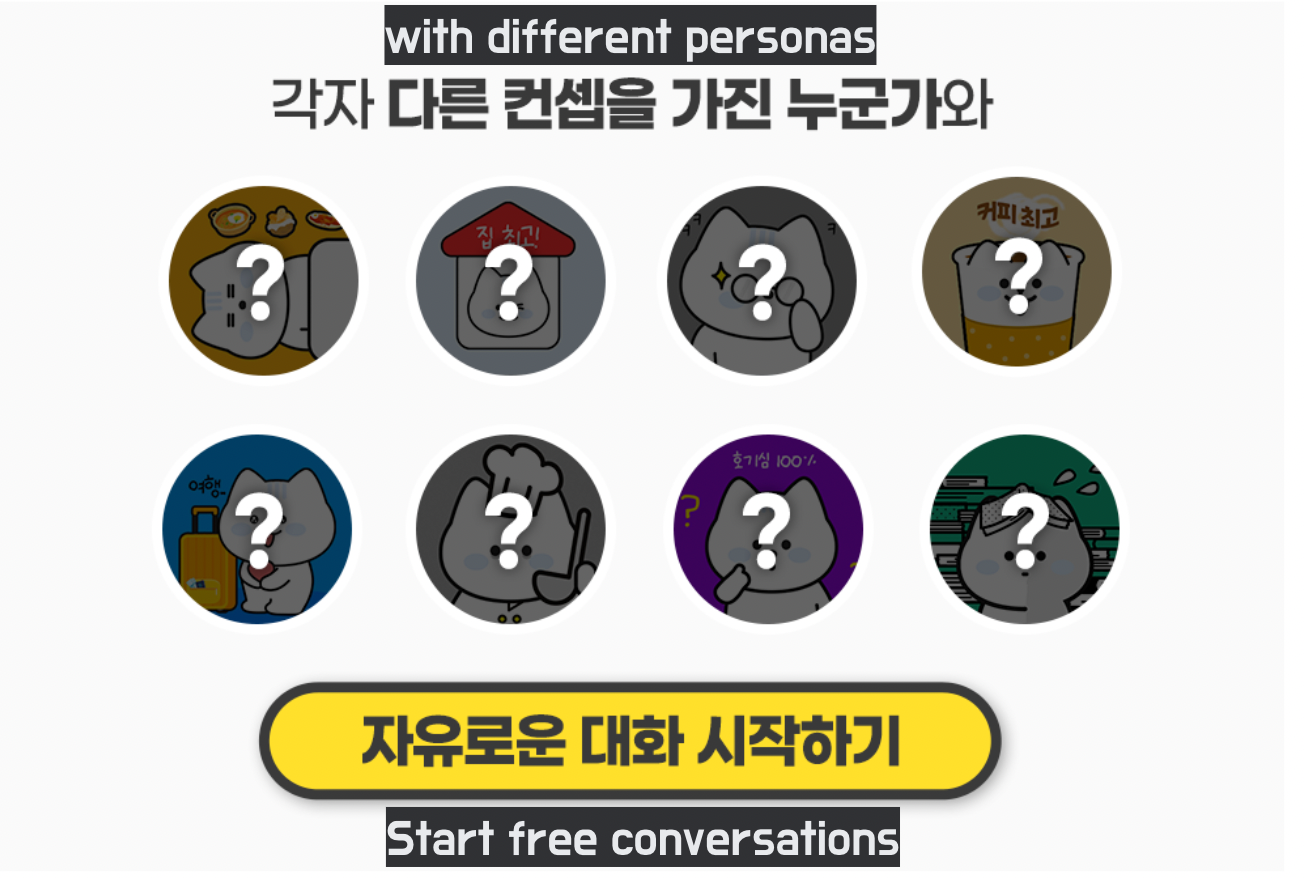} 
\caption{Announcement for the crowd, containing persona profiles.}.
\label{fig:announcement}
\end{figure}

Researchers finally decided on 11 actors after adjudication. Each actor's profile (submitted when they applied) was announced to the worker community to let workers be aware of the project and get interested in each persona (Figure~\ref{fig:announcement}).

\subsection{Dialogue Collection}

In the main collection phase, recruited actors talk with workers in-the-wild. The chatting platform is newly defined, and we introduce several other instructions that fit with large-scale construction, regarding participating time, compensation, etc.

\subsubsection{Platform and Procedure}

\begin{figure}[t]
\centering
\includegraphics[width=\columnwidth]{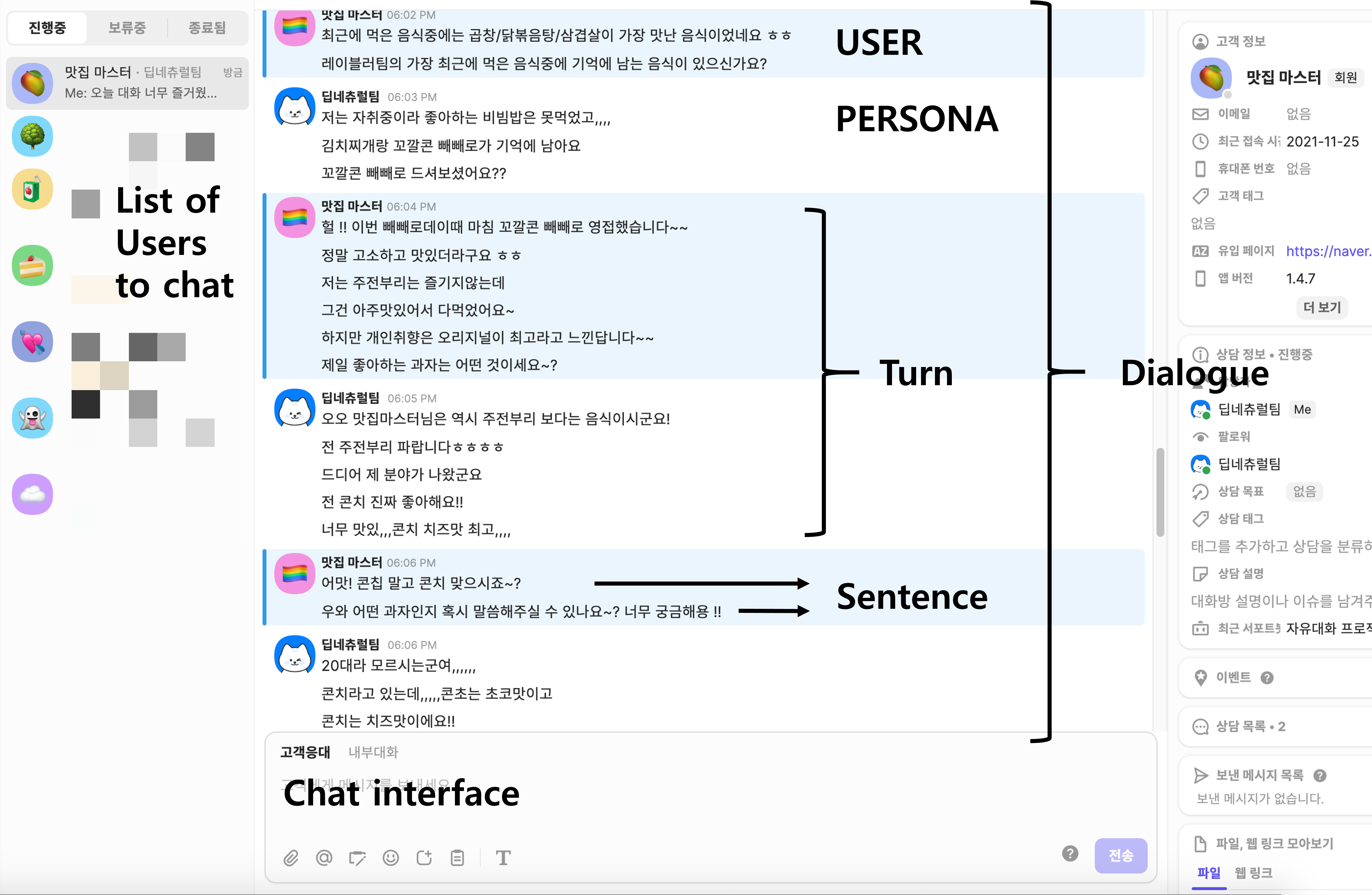} 
\caption{Screenshot of chat user interface.}.
\label{fig:interface}
\end{figure}

We use a user interface that differs from that of the pilot study (Figure~\ref{fig:interface}). It is an application program developed by the crowdsourcing platform, and it has a structure in which workers can open a 1:1 chatting room with an actor. In addition, the moderator can monitor these open rooms, and this monitoring system is used to give penalties to offensive workers or to moderate the overall workload. In detail, \texttt{enter} after typing a sentence makes a line shift, but the turn goes to the counterpart only if the other side starts chatting. The speaker information is exported along with the line change information, and the timestamp is recorded along with the conversation.

Workers read recruitment and participate in projects that are open for each persona. Therefore, the amount of work assigned to each actor is different, and there is no obligation for a worker to have a dialogue with all actors. However, considering the imbalance in the number of participants between actors and users, we informed workers that the actor response could be delayed from time to time. In addition, they were instructed not to push the actor's response, and give a closing remark (e.g., \textit{I am leaving now for dinner!}) at the end of each dialogue so that the actor does not wait for the following conversation.

Basically, the conversation guidelines for persona and user side are similar to that of the pilot study. However, the following cautions were provided to workers to guarantee the consistency and sustainability of the data.

\begin{itemize}
    \item Typos and errata, or non-grammatical sentences that can appear naturally during chatting are allowed.
    \item Simple exclamation or fillers such as ㅋㅋ/ㅎㅎ (lol) are usable, but not accepted as an effective sentence input.
    \item Avoid using photos, emojis, emoticons, etc., since they cannot be shown consistently in multiple data formats.
\end{itemize}

From now on, we introduce some other features added to accommodate conversation flows in the crowdsourcing condition.

\begin{figure*}[t]
\centering
\includegraphics[width=0.85\textwidth]{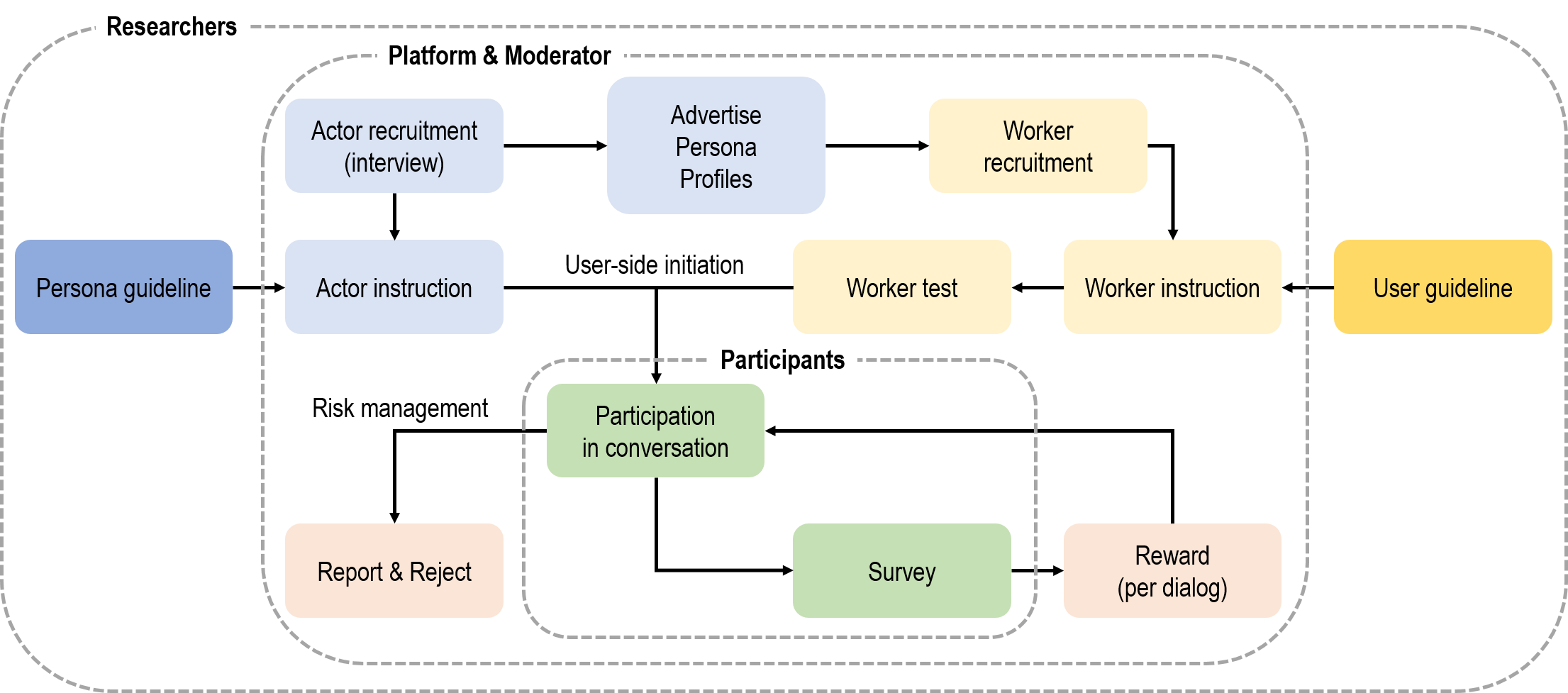} 
\caption{The overall flow of the proposed construction method. Here, \textit{actors} denote participants regarding \textit{persona}, and \textit{workers} indicate who perform \textit{user}. Specific setting of the profile is required only for actors, while conversations are initiated by workers.}
\label{fig:flow}
\end{figure*}

\subsubsection{Artificial Duration}

We benchmarked \citet{xu2021beyond} and instructed the actors to insert artificial durations after at least four turns of conversation. Artificial duration can be freely stretched from several hours (1-7) to several days (1-7), and participants should continue a proper conversation after the insertion of the duration.

The purpose of introducing artificial duration is twofold. First, artificial duration enables the generation of long-term conversations that cannot but be limited in the circumstance of dialogue generation in a limited amount of time. Although the insertion is artificial, participants can reproduce the behavior they have learned in past conversations. Next, by allowing the artificial duration of the persona side, we allow actors to give an unwilling conversation a pause. This serves to alleviate the actor fatigue, and can also be interpreted as a kind of warning message on the user side.

We do not force persona actors to synchronize the same dialogue history for every user in regarding that all the dialogues are considered independent events, except for the preservation of the profile information. For example, a persona may disclose the colloquial events or traveling experiences when talking to user A, but do not have to preserve those settings when talking to user B. This is because every conversation takes place in asynchronous time and space. By inserting an artificial duration, we compensate for the limitation of the narrow time window that can be induced in independent scenarios. At this time, for the natural flow of the conversation, we asked the users to continue the conversation taking into account the actual insertion time without being embarrassed by the artificial duration mentioned by the actors.




\subsubsection{Reward System}

In crowdsourcing, actors were guided to finish a certain amount of dialogue (300 per actor) over a specific period (about one and a half months) or be rewarded with a specific compensation per dialogue. Thus, their work is considered a paid job, so their productivity is less influenced by their willingness to talk with others.
However, user participants take part in a conversation given the existence of a proper reward. Therefore, we built an appropriate reward system by strictly defining \textit{dialogues, turns,} and \textit{sentences} (or sentence-like chunks).

\textit{dialogue} consists of 15 to 30 turns of conversation, including whole lines of text which start with a worker's greeting and end with the finishing part of the conversation. It includes at least three times artificial durations.
\textit{Turn} denotes a set of two sentence-like chunks that the worker and the actor utter subsequently. Here, a sentence-like chunk can be comprised of various lines of sentences, while it does not matter if it is a grammatical sentence.
\textit{Sentence} indicates various lines of text that make up a sentence-like chunk. Therefore, a turn finally consists of more than two lines of sentences uttered by the worker and the actor.

Workers can reach the reward by performing five dialogues, and receives about \$1.2 per dialogue after completing a survey form. The survey form is inspired by the one used in the pilot phase, and includes the review on the other side of the conversation. A similar survey is assigned for actors, but the accomplishment of the survey form is not directly related to the reward in the case of actors.

In each dialogue, workers are obliged to conduct at least 15 turns of conversation. As an incentive, after 15 turns of conversation, additional chatting is available for another 15 turns. At this time, workers receive an additional reward of \$0.4 every five turns. Through this, we designed a system that enables workers to receive additional rewards according to their interests or passion while guaranteeing a minimum amount of conversation.

\subsubsection{Survey Form}

Similar to the pilot study, we asked the participants to fill up the survey after each conversation. Here we describe what we wanted to check in the survey form. A poll was conducted so that participants could evaluate one another's attitudes and the content of their interactions. 
On a scale ranging from 1 (``very low") to 4 (``very high"), we measured the fun of the conversation with the other party (fun; "how fun was the conversation with your partner?"), the naturalness of the conversation flow (naturalness; ``How natural was the conversation flow?"), and the engaging personality of the partner (engaging; ``how engaging was your conversation partner?"). In addition, the participants were asked to evaluate one other's empathy. We adopted a measure, Barrett-Lennard Relationship Inventory (BLRI), used by counselors and clients to assess each other's capacity for empathy in counseling \cite{barrett1981empathy,barrett2015relationship}. The BLRI survey included a scale ranging from 1 (``extremely low") to 4 (``very high").

\section{Project Flow}

The overall project flow is shown in Figure~\ref{fig:flow}. Three stakeholders appear namely \textbf{researchers} who make up persona and user conversation guideline, a \textbf{platform (moderator)} that recruits actors and workers and instruct them to conduct persona dialogues, and the \textbf{participants} who are engaged in the construction. Users initiate all the conversations, and some conversations between actors and workers are halted after the report \& reject process; otherwise, the reward is given after the survey.

Before starting the conversation, workers are informed of the instructions and prepare the dialogue by being familiar with the introduction of the persona. The conversation is started by saying hello, and such greetings are recommended to reflect the characteristics of each persona, not just a simple greeting.

The conversation continues over 15 turns. For each turn, which consists of the worker and the actor’s words, one’s sentence continues until the other side chats. Such chatting is considered a single sentence regardless of the number of lines of text (which equals the number of \texttt{enter}s that are pressed).

The worker terminates the conversation if they decide to quit the dialogue. Sometimes the actor terminates the dialogue if they feel fatigued or eeriness from talking with the worker. Both sides have to finish the survey after each dialogue. Workers get rewarded if they complete the survey form. The whole project ends if the actors reach the number of dialogues assigned to each, about 300 per persona.

\section{Interview}

To guarantee the reproducibility using the proposed scheme, we interviewed the moderator of the platform and four actors who played persona (\#1 - \#4). We summarize interviews to get the assessment of the proposed scheme by the participants. The question sets used for the interview and the full text are provided in the Appendix.

\subsection{Actors' Feedbacks}

\paragraph{Characteristics of persona dialogues} Different from usual conversations with friends or family, \#1 felt the obligation to continue the conversation. \#1 and \#2 felt close to talking with newcomers or colleagues rather than friends, and it led to more considerate word selection (\#2) or careful self-introspection (\#4).

\paragraph{Challenges of talking with real users} Actors were burdened by the number of conversations they had to participate in at the same time (\#2, \#3). They had some challenges when they met partners who were unfavorable, rude, or thoughtless, (\#1) and even some were too pessimistic or made actors feel depressed (\#2, \#4). Some users always asked meaningless questions (e.g., \textit{ what are you doing right now?}) (\#2) or kept asking (\#2, \#4), which made actors feel like being inspected (\#4). Some users even ignored rules or did not listen to the notice, and awkwardly continued conversation after the artificial time (\#2).

\paragraph{Opinions on artificial duration} Actors used artificial duration to avoid boring conversation (\#1), to have a break such as a lunchtime (\#2), or to run the conversation in parallel with other work (\#4). Even there was a case where the conversational situations or flow required artificial duration (e.g., going to an art exhibition virtually) (\#3). Some of them wanted smaller number of mandatory insertion (\#2, \#3) and more free usage.

\paragraph{Difficulty of pursuing the persona} Actors had some difficulty in responding to all the same questions from users, based on their profiles from users (\#3, \#4). They had to prepare a lot to maintain their persona concepts (\#1, \#3). \#1 prepared a list of questions and answers and \#3 learned some new knowledge related to their own profiles. \#2 wanted to change one’s persona because its concepts were too extensive and \#4 had trouble finding the balance between the roles.

\paragraph{The effect of the project in real life} Some actors reported they gradually adjusted to the personas’ characteristics they set (\#1, \#2). \#1 sometimes happened to tell even trivial things to friends since they felt like users in the project, and \#4 started to give an immediate answer in real-world conversations as well because of the project. Moreover, they broadened their personal hobbies or artistic tastes thanks to their partners’ recommendations (\#2, \#3).

\subsection{The moderator's Feedbacks}

\paragraph{Challenges of organizing a large-scale persona dialogue collection} The moderator felt the biggest challenge in giving motivation to actors and understanding their emotions. Actors displayed difficulty in performing the conversation with user participants, and such sharing was usually done via direct message to the moderator. The moderator reported that actors especially faced struggle when workers talked to them in a way actors felt annoying but they could not show the inconvenience (to adjust to their persona), and that the moderator had to emotionally empathize with their struggle while giving a practical solution. The moderator also felt challenged when actors did not share their opinion or emotion frankly, making it difficult to discern the internal status of the actors and give a solution.

\paragraph{Tips on actor recruitment and compensation} Actor recruitment process was done by both researchers and the moderator, but interviews were performed by the moderator in our project. The moderator reported difficulty in communication that is experienced when an actor shows a large discrepancy between the prepared persona and the genuine personality, that it might be a solution for better communication to encourage actors to preserve their aspect of self-disclosure or opinion sharing while making up a persona. In compensation, the moderator reported that actors preferred providing a reward per dialogue to being paid for working for a fixed period with a specific quantity as a goal. The main reasons are the pressure of goal qualification in a fixed working setting and the flexibility of time distribution in a per-dialogue setting.

\paragraph{The reproducibility of the project} The moderator reported that a load of the moderator in this project holds about 10\% to 15\%, while the most important role of s/he is bridging the communication between workers and caring for their mental status. This led to the conclusion that the most important virtue of a moderator is the ability to empathize with the inconvenience of workers and provide comfort. Finding an appropriate candidate for moderator and letting them understand the task to fluently communicate with researchers and workers will help the construction project be successfully performed.

\section{Experiments}

After the whole construction of the corpus, we conduct several experiments to check the characteristics of the dataset. First, we see the overall distribution and corpus statistics that can be obtained from the survey results. Next, we see if the content of dialogue regarding each persona reflects their profile, using word vector clustering. Finally, we check if few-shot dialogue generation is available with large language models, which means that the corpus can be utilized in the way we intended. 

\subsection{Survey Analysis}
We analyze the survey results that participants reported after each dialogue. A total of 1,658 survey data were analyzed. We use Spearman's correlation \cite{10.2307/1412159} to measure the correlation between factors in dialogue statistics and survey results. 

The experiences for both sides in the conversation were not the same. In all seven attributes (fun, naturalness, engaging, BLRI, likeableness, respectfulness, and connectedness), the result of the same sentiment that actors and workers reported did not show any significant correlation (i.e., the correlation between the user and the persona's answer about `How fun was the conversation with your partner?': 0.044, p-value: 0.073).
However, there was a high correlation between some of the seven attributes in each one's response (i.e., the correlation between the user's answer about `How fun was the conversation with your partner?' and `How likeable was your partner during the conversation?' : 0.708, p-value: 1.59e-252).

As it was directed to actors to use pause if the conversation did not go smoothly, the more an actor pauses, the more the actor feels negative about the conversation (i.e., the correlation between the number of pauses and naturalness that the persona felt toward the user: -0.228, p-value: 5.35e-21).
It is noteworthy that workers had a positive conversation experience when there were many pauses (i.e., the correlation between the number of pauses and naturalness that the user felt toward the persona: 0.118, p-value: 1.54e-6).

Additionally, we conduct additional analysis using Welch's t-test \cite{welch1947generalization}.
Actors appraised that the partner was more likeable as the worker was younger (p-value: 0.041).
There was a significant age difference between the group that experienced open chatting (75\% of the total) and the group that did not experience it (p-value: 4.75e-27).
Actors felt more engaged with the worker When the worker experienced open chatting (p-value: 0.045).

\subsection{Topic Clustering}
\begin{figure}[htp]
    \centering
    \includegraphics[width=240pt]{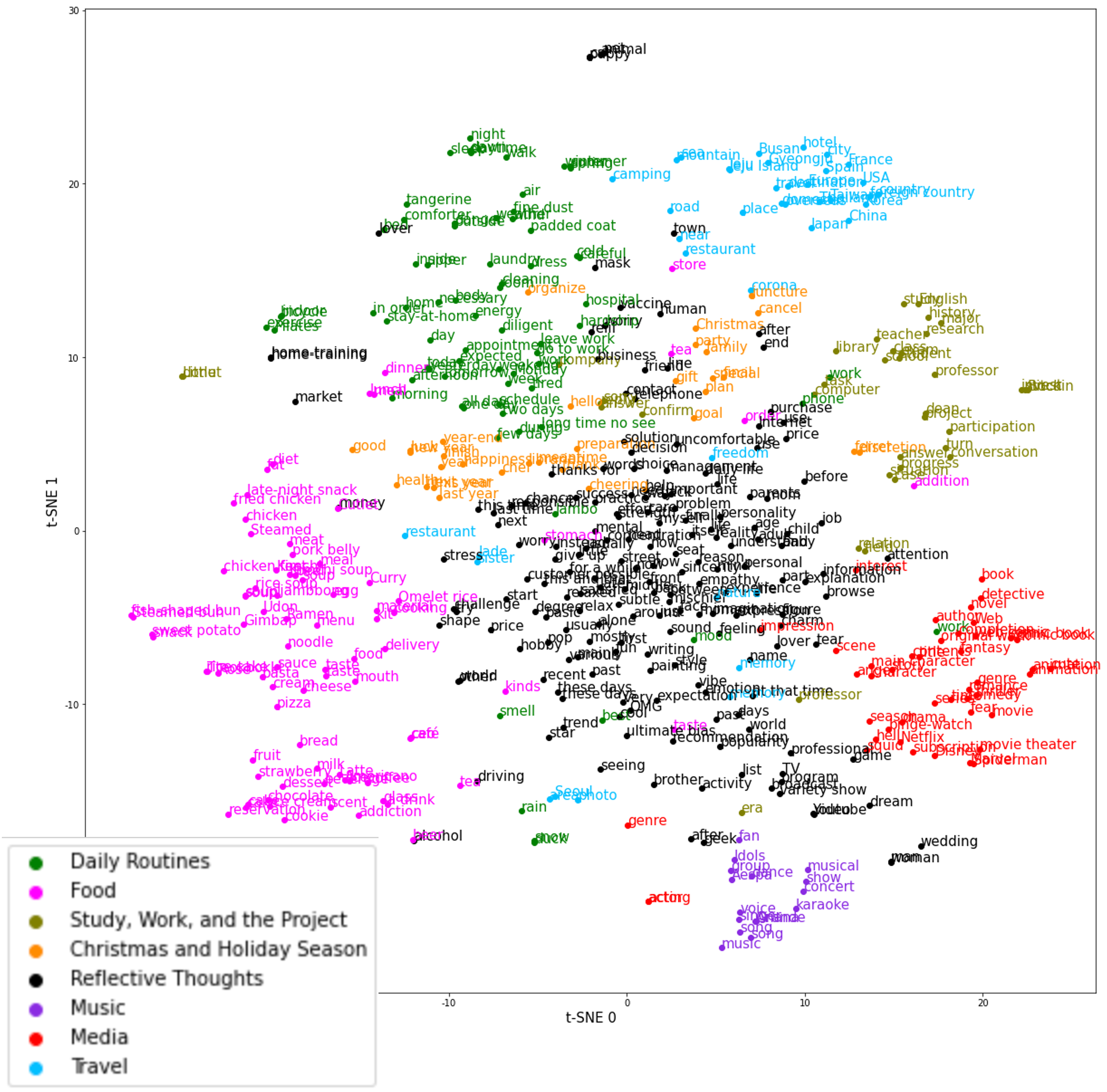}
    \caption{Semantic visualization of words.}
    \label{fig:topic_clustering}
\end{figure}
First, we derived Korean nouns (e.g., common nouns, proper nouns, etc.) using the open-source Python library MeCab tokenizer \cite{kudo2006mecab}. Then we removed some stop words, including dependent nouns, exclamations, and predispositions. We then trained a Word2Vec model \cite{mikolov2013distributed} with Gensim library \cite{rehurek2011gensim} to obtain the final list of tokenized words (N = 452,560). Finally, we created word embeddings with Skip-gram, and the embedded vectors were categorized by K-means clustering.

Figure~\ref{fig:topic_clustering} is the outcome of using t-SNE to classify the embedded vectors \cite{van2008visualizing}. The final top 8 topic clusters are as follows: Daily routines (topic 1), food (topic 2), work/education, or the crowdsourcing project itself (topic 3), Christmas and holiday season (topic 4), reflective thoughts (topic 5), music (topic 6), mass media (topic 7), and travel (topic 8).

\subsection{Few-shot Dialogue Generation}

Beyond simply listing the quantitative characteristics of the collected conversations, we want to show that the resulting corpus can be utilized in these days' few-shot dialogue generation paradigm \cite{brown2020language}. 
In this regard, we select dialogue from the collection and conduct a few-shot generation to see i) if the generated dialogues are aligned with the original script, ii) if adding persona information differs the result, and iii) if the created dataset is useful for future dialogue generation systems. 

Below is the original script of a dialogue conducted between \textit{Hungry sloth} (\texttt{sloth}, persona) and \textit{Green carrot} (\texttt{carrot}, user). Note that the bold part is to be used as an input query for future generation process and the italicized part is the original script that is created by participants.\smallskip

\begin{dialogue}
	\speak{Carrot} Hi I contacted you since you'd be active now. What were you doing Sloth? ... Are you still asleep?
	\speak{Sloth} I'm here carrot how's it going!
	\speak{Carrot} Finally! Thought you were sleeping for 3 days
	\speak{Sloth} Sorry I've been doing an urgent work 
	\speak{Carrot} \textbf{Oh I see.. then the job is now over?}
	\speak{Sloth} \textit{Yes... but I am not in a good condition for being in a different life pattern for three days... I need something to cheer me up}
	\speak{Carrot} \textit{How about taking a deep sleep?}
	\speak{Sloth} \textit{May I sleep from now on? lol} [3 days passed]
\end{dialogue}\smallskip

To see how large language models generate the dialogue with few-shot learning, we adopted Korean GPT-3 API HyperCLOVA\footnote{\url{https://clovastudio.ncloud.com/playground}} \cite{kim2021changes} to design two scenarios\footnote{Though high performance large language models (LLMs) came out after November 2022 presentation of the non-archival version of this manuscript at HCOMP, we did not perform additional experiments with them since the performance regarding Korean text was not guaranteed and some of them readily displayed the self as an intelligent agent, which makes it difficult for us to apply them to our dataset.}. Primarily, we generate further dialogue with the first four sentences of the above conversation, and then compare it with the case of additionally utilizing the persona profile. 
\smallskip

\begin{dialogue}
	\speak{Carrot/Sloth} [Same as above]
	\speak{Carrot} \textbf{Oh I see then the job is now over?}
	\speak{Sloth} \textit{Yes I'm gonna take a rest today}
	\speak{Carrot} \textit{Well done :) working so hard may harm yourself. Btw the weather became so cold these days :(}
	\speak{Sloth} \textit{that's so true... }
\end{dialogue}\smallskip

We noted that the big model gave quite a natural answer sequence. Here, we add the persona profile of Figure~\ref{fig:profile} to the beginning of the dialogue. This gives a slightly different output, which better reflects the persona on the generated dialogue, with just a simple augmentation of the profile text.\smallskip

\begin{dialogue}
    \speak{Profile of Hungry Sloth} Fantasy novel fanatic for over 30 years. Enjoy various genres of comics, webtoons, movies, music, and performances. Night owl. Sticking to hot coffee even in midsummer. Anyone who feeds can be my friend. Get easily depressed when hungry. Usually very positive and optimistic, like a sloth.
	\speak{Carrot/Sloth} [Same as above]
	\speak{Carrot} \textbf{Oh I see then the job is now over?}  \textit{Would you have a dinner with me? It's my treat}
	\speak{Sloth} \textit{Oh that's nice of you!! I am soooo hungry now}
	\speak{Carrot} \textit{Yey see you soon :D}
\end{dialogue}\smallskip

Note that overall responses of Sloth seem to reflect the characteristics of the persona itself. In the original script, taking into account the context that there had been urgent work for a few days, the actor asks for something to cheer up and asks if s/he could sleep, which shows the similarity with Sloth. In the script generated without a persona profile, Sloth takes a rest and the Carrot worries about it, showing the dialogue comparable to the original script. In contrast, the script generated with the profile yields the persona eager to eat something (given that the user treats him/her), highlighting the `hungry' part of the characteristics. 

In generated paragraphs, we checked that the generated dialogues are aligned with the original version, and adding persona information emphasizes another part of the characteristics, inducing a slight change in the dialogue flow. A difference between the original script and the generated ones is that the user-side conversation becomes much more substantial compared to the actor side. This is not an intended change, and it seems that the model places equal or more importance on generating the user-side scripts. However, we think this kind of variety can be studied and exploited positively depending on the nowadays few-shot and prompting paradigm. We believe our dataset can play an essential role in checking the feasibility.

\section{Discussion}

With above settings and collection schemes, we aim to show that our strategy helps make up a large-scale, open-domain persona dialogue corpus with a small group of actors and crowd-user participants, providing both groups with a satisfactory experience in a talk-to-earn paradigm. 

\paragraph{RQ 1:} What should be considered in accommodating the construction of a successful dialogue dataset? Which factors are important for persona dialogues and which make the process challenging?

Organizing persona dialogue differs a lot from usual conversations. Especially when the persona is assigned only to actors (and workers initiate the conversation), it is crucial to handle unexpected and unwanted situations that may make actors embarrassed or annoying. 
It is important to lessen or mediate the mental fatigue within actors, and allowing artificial duration was one solution. The moderator also played an important role here.

\paragraph{RQ 2:} What is the role of the moderator in large-scale dialogue dataset construction? How is this different from other natural language process datasets?

The usual role of moderators in natural language dataset construction is understanding the researchers' tasks and educating workers, while moderating conflicts and managing finance. In other (linguistic) annotation tasks, adjusting to the above role may lead to a successful project. However, the moderator in large-scale dialogue dataset construction may have to be concerned about the conflict between participants since dialogues inevitably involve the interaction between more than two individuals.  Especially in persona dialogue, actors are required to control their social attitude towards the conversational partner, which may result in mental stress. It is the role of the moderator to empathize with the struggles and sometimes give a solution and motivate them, encouraging actors to continue the conversation.

\paragraph{RQ 3:} Will the above considerations help reach an intended construction process and output? Will they guarantee the satisfactory experience of workers, at the same time preserving diverse persona characteristics in the corpus?

First, from topic clustering \cite{van2008visualizing}, we could see that hiring the persona actors with a specific profile and letting them have a dialogue with various users can guarantee diversity of conversation topics. Though there are two main backgrounds; the participants' capability of handling diverse topics, and recruiting actors considering wide coverage of personas, we deemed our approach may prevent the dataset from containing only monolithic conversations. Next, from the few-shot learning example \cite{brown2020language,kim2021changes}, we could check that the persona information can be encoded in the response generation process of the persona side, implying the utility of the constructed dataset in the nowadays prompting paradigm. Turning to worker experience, it was discerned from the survey results that the conversation being fun and the worker being in youth yields the likeableness of the counterpart, and the experience of open chatting leads the counterpart to feel engagingness.  This suggests that the crowdworkers' familiarity with the working environment can influence the atmosphere of the conversation and it might eventually influence the satisfaction. We believe that creating the dialogue is a representative work that requires both workers' cooperation and their satisfactory work experience for high-quality output, and that our conclusion can be beneficial to other crowdsourcing domains with similar struggles.

\section{Conclusion}

In this paper, we have proposed a construction scheme for a large-scale persona dialogue dataset, accompanying the building process, crowdworker interviews, and experiments for validation. We suggested three research questions with our findings, confirming that our approach can answer unanswered areas regarding participants' struggles, moderators' roles, and the output product's appreciation and evaluation. We hope our research can make a footstep to facilitating study on successful and worker-centric corpus building, which can bring satisfactory experience for all the stakeholders that participate in the project. The disclosed version of the collected dialogues are available at the Github respository\footnote{\url{https://github.com/smilegate-ai/OPELA}}, and the material is to be continuously maintained and improved for future research.

\bibliography{aaai22}

\begin{thebibliography}{34}
\providecommand{\natexlab}[1]{#1}

\bibitem[{Barrett-Lennard(1981)}]{barrett1981empathy}
Barrett-Lennard, G.~T. 1981.
\newblock The empathy cycle: Refinement of a nuclear concept.
\newblock \emph{Journal of counseling psychology}, 28(2): 91--100.

\bibitem[{Barrett-Lennard(2015)}]{barrett2015relationship}
Barrett-Lennard, G.~T. 2015.
\newblock \emph{The relationship inventory: A complete resource and guide}.
\newblock John Wiley \& Sons.

\bibitem[{Brown et~al.(2020)Brown, Mann, Ryder, Subbiah, Kaplan, Dhariwal,
  Neelakantan, Shyam, Sastry, Askell et~al.}]{brown2020language}
Brown, T.; Mann, B.; Ryder, N.; Subbiah, M.; Kaplan, J.~D.; Dhariwal, P.;
  Neelakantan, A.; Shyam, P.; Sastry, G.; Askell, A.; et~al. 2020.
\newblock Language models are few-shot learners.
\newblock \emph{Advances in neural information processing systems}, 33:
  1877--1901.

\bibitem[{Byon(2006)}]{Byon+2006+247+276}
Byon, A.~S. 2006.
\newblock The role of linguistic indirectness and honorifics in achieving
  linguistic politeness in Korean requests.
\newblock 2(2): 247--276.

\bibitem[{Cassell, Gill, and Tepper(2007)}]{cassell2007coordination}
Cassell, J.; Gill, A.; and Tepper, P. 2007.
\newblock Coordination in conversation and rapport.
\newblock In \emph{Proceedings of the workshop on Embodied Language
  Processing}, 41--50.

\bibitem[{Davidson et~al.(2017)Davidson, Warmsley, Macy, and
  Weber}]{davidson2017automated}
Davidson, T.; Warmsley, D.; Macy, M.; and Weber, I. 2017.
\newblock Automated hate speech detection and the problem of offensive
  language.
\newblock In \emph{Proceedings of the international AAAI conference on web and
  social media}, volume~11, 512--515.

\bibitem[{Dinan et~al.(2019)Dinan, Roller, Shuster, Fan, Auli, and
  Weston}]{dinan2018wizard}
Dinan, E.; Roller, S.; Shuster, K.; Fan, A.; Auli, M.; and Weston, J. 2019.
\newblock Wizard of Wikipedia: Knowledge-Powered Conversational Agents.
\newblock In \emph{International Conference on Learning Representations}.

\bibitem[{Godfrey, Holliman, and McDaniel(1992)}]{godfrey1992switchboard}
Godfrey, J.~J.; Holliman, E.~C.; and McDaniel, J. 1992.
\newblock SWITCHBOARD: Telephone speech corpus for research and development.
\newblock In \emph{Acoustics, Speech, and Signal Processing, IEEE International
  Conference on}, volume~1, 517--520. IEEE Computer Society.

\bibitem[{Grice(1975)}]{LogicandConversation}
Grice, H.~P. 1975.
\newblock \emph{Logic and Conversation}, 41 -- 58.
\newblock Leiden, The Netherlands: Brill.
\newblock ISBN 9789004368811.

\bibitem[{Kim et~al.(2021)Kim, Kim, Lee, Lee, Kwak, Hyeon, Park, Kim, Kim, Seo
  et~al.}]{kim2021changes}
Kim, B.; Kim, H.; Lee, S.-W.; Lee, G.; Kwak, D.; Hyeon, J.~D.; Park, S.; Kim,
  S.; Kim, S.; Seo, D.; et~al. 2021.
\newblock What Changes Can Large-scale Language Models Bring? Intensive Study
  on HyperCLOVA: Billions-scale Korean Generative Pretrained Transformers.
\newblock In \emph{Proceedings of the 2021 Conference on Empirical Methods in
  Natural Language Processing}, 3405--3424.

\bibitem[{Kudo(2006)}]{kudo2006mecab}
Kudo, T. 2006.
\newblock Mecab: Yet another part-of-speech and morphological analyzer.
\newblock \emph{http://mecab. sourceforge. jp}.

\bibitem[{Li et~al.(2017)Li, Su, Shen, Li, Cao, and Niu}]{li2017dailydialog}
Li, Y.; Su, H.; Shen, X.; Li, W.; Cao, Z.; and Niu, S. 2017.
\newblock DailyDialog: A Manually Labelled Multi-turn Dialogue Dataset.
\newblock In \emph{Proceedings of the Eighth International Joint Conference on
  Natural Language Processing (Volume 1: Long Papers)}, 986--995.

\bibitem[{McTear(2018)}]{mctear2018conversational}
McTear, M. 2018.
\newblock Conversational modelling for chatbots: current approaches and future
  directions.
\newblock \emph{Studientexte zur Sprachkommunikation: Elektronische
  Sprachsignalverarbeitung}, 175--185.

\bibitem[{Mikolov et~al.(2013)Mikolov, Sutskever, Chen, Corrado, and
  Dean}]{mikolov2013distributed}
Mikolov, T.; Sutskever, I.; Chen, K.; Corrado, G.; and Dean, J. 2013.
\newblock Distributed representations of words and phrases and their
  compositionality.
\newblock In \emph{Proceedings of the 26th International Conference on Neural
  Information Processing Systems-Volume 2}, 3111--3119.

\bibitem[{Moon, Cho, and Lee(2020)}]{moon2020beep}
Moon, J.; Cho, W.~I.; and Lee, J. 2020.
\newblock BEEP! Korean Corpus of Online News Comments for Toxic Speech
  Detection.
\newblock In \emph{Proceedings of the Eighth International Workshop on Natural
  Language Processing for Social Media}, 25--31.

\bibitem[{Rashkin et~al.(2019)Rashkin, Smith, Li, and
  Boureau}]{rashkin2019towards}
Rashkin, H.; Smith, E.~M.; Li, M.; and Boureau, Y.-L. 2019.
\newblock Towards Empathetic Open-domain Conversation Models: A New Benchmark
  and Dataset.
\newblock In \emph{Proceedings of the 57th Annual Meeting of the Association
  for Computational Linguistics}, 5370--5381.

\bibitem[{Rehurek and Sojka(2011)}]{rehurek2011gensim}
Rehurek, R.; and Sojka, P. 2011.
\newblock Gensim--python framework for vector space modelling.
\newblock \emph{NLP Centre, Faculty of Informatics, Masaryk University, Brno,
  Czech Republic}, 3(2): 2.

\bibitem[{Roller et~al.(2020)Roller, Boureau, Weston, Bordes, Dinan, Fan,
  Gunning, Ju, Li, Poff et~al.}]{roller2020open}
Roller, S.; Boureau, Y.-L.; Weston, J.; Bordes, A.; Dinan, E.; Fan, A.;
  Gunning, D.; Ju, D.; Li, M.; Poff, S.; et~al. 2020.
\newblock Open-domain conversational agents: Current progress, open problems,
  and future directions.
\newblock \emph{arXiv preprint arXiv:2006.12442}.

\bibitem[{Roller et~al.(2021)Roller, Dinan, Goyal, Ju, Williamson, Liu, Xu,
  Ott, Smith, Boureau et~al.}]{roller2021recipes}
Roller, S.; Dinan, E.; Goyal, N.; Ju, D.; Williamson, M.; Liu, Y.; Xu, J.; Ott,
  M.; Smith, E.~M.; Boureau, Y.-L.; et~al. 2021.
\newblock Recipes for Building an Open-Domain Chatbot.
\newblock In \emph{Proceedings of the 16th Conference of the European Chapter
  of the Association for Computational Linguistics: Main Volume}, 300--325.

\bibitem[{Schuetzler et~al.(2018)Schuetzler, Giboney, Grimes, and
  Nunamaker~Jr}]{schuetzler2018influence}
Schuetzler, R.~M.; Giboney, J.~S.; Grimes, G.~M.; and Nunamaker~Jr, J.~F. 2018.
\newblock The influence of conversational agent embodiment and conversational
  relevance on socially desirable responding.
\newblock \emph{Decision Support Systems}, 114: 94--102.

\bibitem[{Seeger, Pfeiffer, and Heinzl(2021)}]{seeger2021texting}
Seeger, A.-M.; Pfeiffer, J.; and Heinzl, A. 2021.
\newblock Texting with humanlike conversational agents: designing for
  anthropomorphism.
\newblock \emph{Journal of the Association for Information Systems}, 22(4): 8.

\bibitem[{Shah et~al.(2018)Shah, Hakkani-T{\"u}r, T{\"u}r, Rastogi, Bapna,
  Nayak, and Heck}]{shah2018building}
Shah, P.; Hakkani-T{\"u}r, D.; T{\"u}r, G.; Rastogi, A.; Bapna, A.; Nayak, N.;
  and Heck, L. 2018.
\newblock Building a conversational agent overnight with dialogue self-play.
\newblock \emph{arXiv preprint arXiv:1801.04871}.

\bibitem[{Shuster et~al.(2021)Shuster, Urbanek, Szlam, and
  Weston}]{shuster2021me}
Shuster, K.; Urbanek, J.; Szlam, A.; and Weston, J. 2021.
\newblock Am i me or you? state-of-the-art dialogue models cannot maintain an
  identity.
\newblock \emph{arXiv preprint arXiv:2112.05843}.

\bibitem[{Smith et~al.(2020)Smith, Bishop, Dambha-Miller, Ratnapalan, Lyness,
  Vennik, Hughes, Bostock, Morrison, Mallen et~al.}]{smith2020improving}
Smith, K.~A.; Bishop, F.~L.; Dambha-Miller, H.; Ratnapalan, M.; Lyness, E.;
  Vennik, J.; Hughes, S.; Bostock, J.; Morrison, L.; Mallen, C.; et~al. 2020.
\newblock Improving empathy in healthcare consultations—a secondary analysis
  of interventions.
\newblock \emph{Journal of general internal medicine}, 35(10): 3007--3014.

\bibitem[{Spearman(1904)}]{10.2307/1412159}
Spearman, C. 1904.
\newblock The Proof and Measurement of Association between Two Things.
\newblock \emph{The American Journal of Psychology}, 15(1): 72--101.

\bibitem[{Stalnaker(2002)}]{stalnaker2002common}
Stalnaker, R. 2002.
\newblock Common ground.
\newblock \emph{Linguistics and philosophy}, 25(5/6): 701--721.

\bibitem[{Tannen et~al.(2005)}]{tannen2005conversational}
Tannen, D.; et~al. 2005.
\newblock \emph{Conversational style: Analyzing talk among friends}.
\newblock Oxford University Press.

\bibitem[{Van~der Maaten and Hinton(2008)}]{van2008visualizing}
Van~der Maaten, L.; and Hinton, G. 2008.
\newblock Visualizing data using t-SNE.
\newblock \emph{Journal of machine learning research}, 9(11).

\bibitem[{Waseem and Hovy(2016)}]{waseem2016hateful}
Waseem, Z.; and Hovy, D. 2016.
\newblock Hateful symbols or hateful people? predictive features for hate
  speech detection on twitter.
\newblock In \emph{Proceedings of the NAACL student research workshop}, 88--93.

\bibitem[{Welch(1947)}]{welch1947generalization}
Welch, B.~L. 1947.
\newblock The generalization of ‘STUDENT'S’problem when several different
  population variances are involved.
\newblock \emph{Biometrika}, 34(1-2): 28--35.

\bibitem[{Wen et~al.(2017)Wen, Vandyke, Mrk{\v{s}}i{\'c}, Gasic, Barahona, Su,
  Ultes, and Young}]{wen2017network}
Wen, T.-H.; Vandyke, D.; Mrk{\v{s}}i{\'c}, N.; Gasic, M.; Barahona, L. M.~R.;
  Su, P.-H.; Ultes, S.; and Young, S. 2017.
\newblock A Network-based End-to-End Trainable Task-oriented Dialogue System.
\newblock In \emph{Proceedings of the 15th Conference of the European Chapter
  of the Association for Computational Linguistics: Volume 1, Long Papers},
  438--449.

\bibitem[{Xu, Szlam, and Weston(2021)}]{xu2021beyond}
Xu, J.; Szlam, A.; and Weston, J. 2021.
\newblock Beyond goldfish memory: Long-term open-domain conversation.
\newblock \emph{arXiv preprint arXiv:2107.07567}.

\bibitem[{Yang, Jang, and Cho(2022)}]{yang2022apeach}
Yang, K.; Jang, W.; and Cho, W.~I. 2022.
\newblock APEACH: Attacking Pejorative Expressions with Analysis on
  Crowd-Generated Hate Speech Evaluation Datasets.
\newblock \emph{arXiv preprint arXiv:2202.12459}.

\bibitem[{Zhang et~al.(2018)Zhang, Dinan, Urbanek, Szlam, Kiela, and
  Weston}]{zhang2018personalizing}
Zhang, S.; Dinan, E.; Urbanek, J.; Szlam, A.; Kiela, D.; and Weston, J. 2018.
\newblock Personalizing Dialogue Agents: I have a dog, do you have pets too?
\newblock In \emph{Proceedings of the 56th Annual Meeting of the Association
  for Computational Linguistics (Volume 1: Long Papers)}, 2204--2213.

\end{thebibliography}

\section{Acknowledgments}
This work is supported by Smilegate AI\footnote{\url{https://smilegate.ai/}}. We thank all our crowdworkers and DeepNatural AI\footnote{\url{https://deepnatural.ai}} for creating high-quality data. 
\clearpage

\appendix

\section{Full Interview Texts}
\label{app:interviews}

\subsection{Question Set}

\subsubsection{Questions for the moderator}

\begin{enumerate}
    \item Had you ever had experience in managing the construction of large-scale dialogue corpus? If not, have you managed creating any natural language dataset?
    \item What was the most challenging point in the process of large-scale dialogue dataset construction? And which point made you worthwhile?
    \item What was the main struggle of actors and how did you cope with such circumstances?
    \begin{enumerate}[label=(\alph*)]
        \item What is the key of the communication with actors?
    \end{enumerate}
    \item Were there any cases that you felt stuck in communicating with the actors? 
    \begin{enumerate}[label=(\alph*)]
        \item Which part made you feel difficulty?
        \item Is such difficulty aligned with any characteristics of persona actors that you observed in interviewing them?
        \item Is there a condition that the communication between the moderator and actors can be improved?
    \end{enumerate}
    \item How was the compensation of the actors decided? Were they satisfied with the reward?
    \item How do you think the significance of the role of the moderator in this construction project?
     \begin{enumerate}[label=(\alph*)]
        \item Which part do you think most important among the role of the moderator?
        \item Were there any points of the project that could be improved?
    \end{enumerate}
\end{enumerate}

\subsubsection{Questions for actors}

\begin{enumerate}
    \item Are you experienced with anonymous chatting before?
    \item What is the main difference of this kind of conversation with usual dialogue with surrounding people?
    \begin{enumerate}[label=(\alph*)]
        \item Which kind of conversation was the closest with this project? (among conversation with friends, colleagues, parents, etc.)
    \end{enumerate}
    \item Was dialogue with real users different from the practice done within an interview?
    \begin{enumerate}[label=(\alph*)]
        \item Was there difference between conversation with a fixed, expected counterpart and an unspecified and probably unexpected crowd?
        \item Was there any strategy of one's own to lead the conversation fluently?
    \end{enumerate}
    \item How is your opinion about inserting artificial duration in between the conversation?
     \begin{enumerate}[label=(\alph*)]
        \item In what situation did you frequently utilize artificial durations?
        \item Did the aspect of using artificial duration differ by the counterpart? If so, in what aspect?
        \item Was it possible to continue natural conversation after inserting artificial duration?
    \end{enumerate}
    \item Other difficulties in the project? Or was there any moment you wanted to quite the conversation?
     \begin{enumerate}[label=(\alph*)]
        \item If so, what was the reason?
        \item Which point was the most difficult in pursuing the persona? Have you struggled anything to preserve the persona? If the profile contained much information about the character, was it memorizable through the dialogues? How much were you aware of your persona while in the dialogue?
        \item Have you ever wanted to change the concept of yourself throughout the project? Was there any other actor you wanted to refer to?
    \end{enumerate}
     \item Other questions on specific dialogues in the project
     \begin{enumerate}[label=(\alph*)]
        \item Who was the most impressive user and  what was the reason? Which characteristic did the conversation have?
        \item Which was the type of user you most wanted to keep the conversation with? Was there anyone you thought as `friend' and if so, why?
        \item Which type of user did you most wanted to stop the conversation with? In other words, in which conversation did you feel far from or awkward with the counterpart?
    \end{enumerate}
    \item How did the project change yourself?
     \begin{enumerate}[label=(\alph*)]
        \item How is the persona similar with you in reality?
        \item Is there any change of yourself in reality before and after the project?
    \end{enumerate}
\end{enumerate}

\subsection{Full Text}

\paragraph{Moderator}
\hfill

\begin{enumerate}

\item \textit{I haven't experienced this large-scale dialogue dataset, even adopting personas. Still, I've had quite a few experiences collecting and managing natural language datasets.}

\item \textit{The most challenging part was giving motivation to actors and understanding their emotions, especially the struggles in conducting the conversations. The process resembles a psychiatric consultation. Thus, I often asked participants if they were okay, for instance, at the beginning of each week. Some actors asked for help if they felt inconvenient or stuck, then I deemed that other actors would think similarly and asked their opinion. I felt most worthwhile when actors were encouraged by my support and consultation and decided to continue their conversation with user participants.}

\item \textit{Actors told me they usually struggled when user participants talked to them in a way they felt annoying, but they could not show any inconvenience (mostly due to their persona). Since there were such frequent cases of such wild users, small actions became the stress trigger as time went on. The remote talking situation also made them difficult to handle their emotion, which sometimes led to the participant's comfort. In this situation, I immediately showed an empathetic stance or resolved the problem if I could, and sometimes this led to the comfort of the participant. Once the issue was resolved, actors felt much more comfortable on that topic in later conversations, and I also added such exceptional cases to the guideline (though not all the problems were 100\% resolved).\\
(a) The key to communication with actors is trust and rapport. Interviewing actors and pre-educating the actor guideline helps make the rapport, and here, talking to them with messengers lets them feel comfort and familiarity with the project itself, aside from an official situation with a contract.I played the main role, thereby communicating with them continuously, and this may have given them trust in me that even allowing them to tell me all the struggles and difficulties they faced in performing the task. In other natural language dataset tasks, especially linguistic ones such as POS tagging, this process may not be required since the task is pre-defined and the workers are mostly experts on that topic, the goal is apparent, and the construction pipeline is processed without difficulties. However, various exceptional conflicts can occur in dialogue dataset creation because the task requires at least two participants whose backgrounds, thoughts, and behavior may differ. This calls for the importance of letting the participants have a responsibility for the project and allowing them an experience of building a rapport with somebody new.}

\item \textit{(a) I felt stuck when I could not discern the actual status of the actors. In this sense, I felt the actors most cooperative when they told me about their struggles in detail or answered frankly to my questions. In contrast, when they say only okay on their status or do not make a choice between multiple candidates, it makes me difficult to understand their circumstances truly.\\
(b) I experienced difficulty when actors did not transparently open their struggle to me, and it was more significant when the persona they've chosen is not aligned with their real-world self. For instance, if their persona is delightful and active while their original personality is calm and quiet, the communication cannot be processed expectedly, preventing the formation of rapport between the actor and the moderator.\\
(c) Acting well and showing a persona completely different from one's genuine personality is undoubtedly a desirable ability of the actors. However, it is a dilemma for the moderator that organizing the communication with such actors (and predicting it during the interview) is quite challenging. Thus, it would be nice if actors set up the persona while preserving their personality on self-disclosure or opinion sharing to prevent the situation that the moderator fails to form a rapport with the actor owing to the double-faced attitude.}

\item \textit{The compensation for the actors was set differently from the user participants. There were two ways of providing the reward: rewarding for a fixed period and rewarding per dialogue. We first chose the former strategy, which aims to complete 600 dialogues per actor, and it lasted about a month. Afterward, after collecting the feedback from actors, the compensation was changed to providing the reward per dialogue. Actors said that the latter is better since rewarding for a fixed period gives them stress for not qualifying the required quantity, and rewarding per dialogue lets them use their time more flexibly.}

\item \textit{(a) I believe the moderator contributes about 10\% to 15\% of the total load in this project. The researchers initiate the project, and the workers do the sourcing, while the moderator is in charge of motivating workers, resolving conflicts between workers or between researchers and workers, and facilitating overall communication (much like a platelet in our blood!). For this, the moderator should be able to empathize with the inconvenience of workers and provide comfort to them, which fits with a dataset construction process like this which requires the formation of rapport with workers.\\
(b) It would have been nice if there was a forum where the persona actors could display their emotions and share the information, which would be better than personally asking the help of the moderator every time. Since the construction was processed remotely, there was hardly a chance that actors could get close to each other, leading to less public disclosure of their difficulty to other actors. If such circumstances can be shared and actors empathize with each other, they will reach a wider variety of viewpoints and solutions than merely relying on the emotional empathy of the moderator.}

\end{enumerate}

\paragraph{Interviewee \#1}

\hfill

\begin{enumerate}
    
\item \textit{No anonymous chat experience before. }

\item \textit{Felt obligation to continue conversation. Felt close to talking with newcomers rather than with friends or family.}

\item \textit{The interviewer (the moderator) was favorable, but the unspecified crowd was not. Some even treated their counterpart as if they were AI. For embarrassing circumstances, I prepared a list of common questions and answers based on my profile (e.g., moviegoing).}

\item \textit{The artificial duration was inserted when the conversation was boring, or a single topic lasted too long; difficult to continue the talking. A few-hour duration was mainly used to avoid confusion, not the exceptionally long case such as 4-5 days later. Some users felt offended because they thought their words were ended abruptly.}

\item \textit{As a person with somewhat strict criteria for my own life, it was challenging to talk with many people who sometimes become rude or thoughtless. I think I became slightly aggressive if my counterpart crossed the line, especially those who mistreated me regardless of my kind and tender persona.}

\item \textit{The most impressive one was when the other told me that s/he was reading the same book I was reading then. We experienced high-level empathy, and the conversation flow went very fluent. The conversation I wanted to stop or quit was when the counterpart talked very subjectively or only told their personal stories without listening to me. Such cases lowered my energy.}

\item \textit{At first, I tried to adjust to my persona and be polite, for example, using honorifics. Still, as the conversation continued, I fit with the counterpart rather than strictly adjusting to the original setting. In real life, I sometimes tell even small things to friends because they feel like users in the project. I usually talk a lot with people around me. Still, the number of conversation partners suddenly soaring up was a challenging experience.}
\end{enumerate}

\paragraph{Interviewee \#2}
\hfill

\begin{enumerate}
\item \textit{Yes, it is similar to how I chat with users in the game I play often.}
\item \textit{The conversation style was similar to how I would converse with my coworkers rather than close friends. I paid more attention to the other person's expression than mine, so I carefully chose words.}
\item \textit{(a) Practicing with the project manager was more engaging and fast-paced. I felt both of us were significantly participating. However, the conversation with crowd workers was slightly different from what I had expected. Some conversations were very slow-paced. Several people would not respond for more than 2 to 3 days. Even I sometimes forgot to check the other person's response, which cut the conversation flow.\\
(b) At first, it took some time for me to adjust to the concept (persona) I created, but most of the crowd workers asked me many questions. Many asked similar questions like "what's your interest? Or what are you most interested in?". In the beginning, I passively answered but later on, to give some fun to the process, I tricked some answers or delved into one idea and kept talking about it.}
\item \textit{(a) The good thing about setting the artificial time was that I could use it freely to have a break, lunchtime, or leave work. Some participants did not want to cut it off during the conversation, so I sometimes used the artificial time to end a conversation with those people peacefully. However, it was a bit awkward when I had to fill the minimum number of times to insert in the middle of the conversation. I thought it would be nice if the minimum number of times were a little less.}
\item \textit{(a) The most challenging part was answering a series of questions about my interests. It wasn't easy because I was constantly asked questions about new interests. It was hard to keep making new interests I haven't mentioned previously. For example, if the partner asked, "what are you interested in?" Then I would answer "electronic devices." A follow-up question would be, "I like Samsung, how about you?" but they refused to follow up and kept asking new questions like, "oh yes, what are your other interests then?". Next, when various interests came up, I searched for them in that situation to make sure I was saying the right thing. As I continued doing this simultaneously, the conversation continued to cut off, and the time was extended a bit. \\
(b) My persona was similar to my authentic self, so I hadn't had trouble memorizing the concept while conversing. \\
(c) I wanted to change my persona every moment during the project. I did check on other personas, but I don't particularly recall which one was the most impressive. But if there is next time, it would be better to advise personas not to make the list of concepts extensive.} 
\item \textit{(a) Two people just come to mind: one person felt excessively close to me (I may be the only one feeling it). It was a fun conversation, though. I don't know why people feel so close to me, but I felt sorry that the conversation had to end like that. \\
(b) First is someone who has a mutual interest. The second is someone who reacts well to what I say. The third is actively participating in the conversation. I could tell whether a person is being pretentious or being sincere just by reading what they say. It seemed obvious that when one is interested in asking and answering questions, even if they are not interested. Some people change the subject no matter what they say to kill time. \\
(c) Although the participant must have already been aware that we should use artificial time and continue talking, some people ignored the rule. They just said what they wanted to say. For example, some people awkwardly continued the conversation after artificial time by suddenly speaking according to the current time. Some particular people said just one word a day. Also, I didn't like people who confused me about whether they wanted to talk to me or ask questions. No matter what you say, you have to keep talking about A and keep talking about A-1, but B keeps coming up, and if you ask B, C comes up. \\
Another challenge was the people who had conversations that made me feel depressed. He was a difficult conversation partner. No matter what these people say, they go into a bad mood. This person didn't say depressing words throughout the conversation, but overall, the sentiment was negative. I am not sure, but I'm guessing it's because he's had bad experiences with friends in the past. I think it was because my persona looked like a so-called 'insider,' like someone extrovert interested in various things.Maybe that's why he wanted me to advise him on keeping a good relationship. I actively gave some tips and guidance, but only negative conclusions were reached no matter what.} \item \textit{(a) When I first started a conversation, I think my authentic self and this persona were about 4-50 percent similar. After several conversations, the similarity reached nearly 60 percent. The difference of 10 percent had changed little by little as I participate more. In other words, I gradually adjusted my habits to the persona I set. For example, I don't go out often, but there was a day when my conversation partner asked, 'Did you go for a walk today?' I didn't go for a walk, but if I said I did, I felt like I was deceiving the other person, so I went out for a walk on purpose, and I made minor changes to my daily life.\\ 
(b) At first, I was burdened by the number of conversations and the constant ringing of the alarm. Coping with this was difficult as many partners ignored the rules or did not listen to the notice again.\\ The positive change was that, as we continued to talk, I was not interested in the fact that the scope of my hobbies was forcibly expanded if the other person was interested. For example, I'm not a fan of American TV series. Still, I got a lot of recommendations from conversation partners, so I learned a lot and many exciting things.\\ Overall, this project was so much fun that I wanted to participate again in the future. I think it's a significant experience. When in my life will I be able to have conversations with people with diverse thoughts and tastes? It was an excellent experience. Such occasions are rare. I'm at work right now, and since this experience, I have talked much with people around me. Then my colleagues said they wanted to try it and even asked how to participate. People around me were also very interested.}
\end{enumerate}

\paragraph{Interviewee \#3} 

\hfill

\begin{enumerate}
\item \textit{As the format of the interview slightly changed depending on the interviewee, we did not ask for the question 1 to the interviewee 2.}
\item \textit{As the format of the interview slightly changed depending on interviewee, we did not ask for the question 2 to the interviewee 2.}
\item \textit{It was a tough time since there was a huge rush of users entering the chatroom at first, and I could not make any conversation as I wanted. But I started to adapt to this situation after I completed a conversation with 20-30 users. I also had some tactics to lead the conversation fluently – bring various interesting topics, no insincere reaction or answer, be natural as if users were my friends, keep thinking and learning new topics or issues, be connected to partners such as empathizing/understanding/compliment}
\item \textit{As there were rules to put artificial duration at least three times, I used to put them though I did not want to. But sometimes I could not help putting them due to some situations in the conversations such as going on a diet or going to art exhibitions, etc. Inserting the artificial durations sometimes made me forget the original flow of the conversations, so using those might not be helpful. It will be much better if the rule is not mandatory.}
\item \textit{ Since I had to change my persona details even after the project started, I had a hard time pretending as if I were a real painter. To solve this problem, I tried to get some new information about art. However, as users watched my profile and initiated conversations, they all came up with and asked the same questions. Even some users who made conversations insincerely always asked me meaningless questions, which irritated me. As I had to answer for many users simultaneously, I could not remember the details of the conversations or sometimes missed the flow.}
\item \textit{Sincere reactions such as "Awesome! You are a genuine painter", "You're very talented! Why don't you start a YouTube channel?", or "It was really glad to meet and talk with you!", were the most memorable comments to me. 
One partner told me that (s) he'd changed a lot after (s)he talked with me, which gave me the power to keep carrying on though the project was somewhat exhausting. 
Those who filled conversations with pointless sentences such as "what are you doing right now?", "did you have lunch?" made me feel tired, and I did not want to answer these kinds of questions.}
\item \textit{Whenever users recommended me some movies or artworks during the conversations, I did watch or appreciate them. That could be some changes in my life after the project. The persona was about 60 percent similar to my real identity. The rest, 40 percent, was because, in some points of view, it reflected my ideal self-image (or it could be a wish or dream).}
\end{enumerate}

\paragraph{Interviewee \#4}
\hfill

\begin{enumerate}
    
\item \textit{Only some random chatting service experiences. Done chatting in some games but not as a different persona}

\item \textit{As this is the first experience, I've projected a part of myself so as not to make a mistake, but sometimes found myself not talking in the way that I intended.}

\item \textit{The crowd's background was diverse and unspecified, so they understood my persona differently.  I thought they would act in some way considering my persona but encountered many exceptional cases, making the answering troublesome. To cope with these cases, I remembered some reactions shown by user participants that seemed similar to me and used them in another conversation.}

\item \textit{As I did this as a part-time job while doing other work, artificial duration was used when I could not respond immediately. Some users reacted appropriately, while others found it difficult to answer properly after the insertion and ignored the artificial duration. It was also challenging to become accustomed to artificial duration (for instance, when the artificial duration is three days but only a half-day in the real world). So I looked up the dialogue history of making an appropriate answer and compensating for the artificial time passed. I did not use them to avoid unfamiliar topics. Using the artificial duration didn't matter much in the overall project, and I frequently used `hour'-level duration since they fit with circumstances when I could not respond for a while when working.}

\item \textit{(a) The most challenging part was the considerable number of participants. It wasn't easy to match the dialogue content with the user, so looking up dialogue history was necessary when making a response. Also, when surveying the conversation, answering if I felt connectedness or fun was a bit tough since the feeling I gen when talking with newcomers is not that diverse. Subjective questions were tough to answer.\\
(b) Persona I intended overlaps about 5-60\% with my original personality (which made me quite considerate in talking with the users). I thought it would benefit the continuation of the conversation. The overlap is disliking reading books, which is intended to induce users' interest. People often asked why I became a librarian, nevertheless not liking books. It was successful at starting the conversation, but responding with the same answer to all made me inconvenient.\\
(c)  People asked for book recommendations since I was a 'librarian' persona, but it was challenging to find the balance between the roles as I positioned mine to dislike reading. Also, as I stated that I like being at home, I had to explain the background for going out, such as meeting friends. I referred to 2-3 other persona actors and found that `Jade' leads conversation fluently while at the same time preserving the persona. I thought that persona would be more sustainable for me.}

\item \textit{(a) The most impressive one was the conversation with `Labeler tired from labeling', who utilized their common sense and knowledge properly in talking about  society. We even talked about the innovation of libraries in the digital era and how the future libraries will be. I was first curious if this kind of conversation fits with the project, but soon became interested in the conversation and asked many things, feeling the fulfillment of the knowledge. Another impressive participant seemed to work as a part-time library worker. Overall, the content of the conversation made me remember the dialogue. \\
(b) I sometimes felt fun in having the conversation. I felt it when they listened sincerely to my opinions, and friendly chat also mattered. In content, daily life was the main topic. Unfortunately, they also terminated the conversation as they reached the aimed amount, though I wanted to talk more with them. \\
(c) I remember two participants. One was too pessimistic, such as `I will be alone on holidays. I am so lonely.' I wanted to end the conversation as fast as I could. I did not know how to handle and care for their emotion. Another participant kept asking something. As I give them the answer, the person brings up another topic, making me feel like being inspected.}

\item \textit{I used to forget to answer the messenger for a few hours, but as I participated in the project, I tried to give an immediate answer in real-world conversation. I had trouble before with this characteristic, but it seems that the project changed my personality.}
\end{enumerate}

\end{document}